# Real-World Modeling of a Pathfinding Robot Using Robot Operating System (ROS)

Sayyed Jaffar Ali Raza[(1)], Nitish A. Gupta[(2)], Nisarg Chitaliya[(3)], Gita R. Sukthankar[(4)]

Department of Electrical Engineering & Computer Engineering
University of Central Florida
32816 Orlando, Florida USA

jaffar@knights.ucf.edu, nitish.gupta@knights.ucf.edu, chitaliya.nisarg4@knights.ucf.edu, gitars@eecs.ucf.edu

## ABSTRACT
This paper presents a practical approach towards implementing pathfinding algorithms on real-world and low-cost non-commercial hardware platforms. While using robotics simulation platforms as a test-bed for our algorithms we easily overlook real-world exogenous problems that are developed by external factors. Such problems involve robot wheel slips, asynchronous motors, abnormal sensory data or unstable power sources. The real-world dynamics tend to be very painful even for executing simple algorithms like a Wavefront planner or A-star search. This paper addresses designing techniques that tend to be robust as well as reusable for any hardware platforms; covering problems like controlling asynchronous drives, odometry offset issues and handling abnormal sensory feedback. The algorithm implementation medium and hardware design tools have been kept general in order to present our work as a serving platform for future researchers and robotics enthusiast working in the field of path planning robotics.

## 1. INTRODUCTION
Robots are one of the most attractive machines in the field of artificial intelligence. They are changing the world with their robust and intelligent behaviors, especially in the industrial sector. Most common type of robot we see is a vehicular shaped robot on wheels equipped with a bunch of sensors and with flexible motion capabilities. It looks very fascinating to see a robot cruising steadily on the ground, avoiding obstacles and reaching from one point to another like a champion. These tasks might look very easy to be implemented conceptually or theoretically, however they involve exponential complexities behind the scene. We can generalize the big-picture of robot building into two sections; algorithm design and physical dynamics. Algorithm design comprises of designing a mathematical model of the world and formulating logical methods for a specific task like collision avoidance, mapping, path planning and finding, localization etc. We will discuss path-finding algorithms and their implementation in next chapters. The part of the big-picture is physical dynamics. It involves modeling of parameters that are directly related to stochastic world behavior. Since algorithms for robot agents are programmed in a computer-generated simulation world and despite those simulation worlds have a stochastic conception, the real-world behavior of the agent differs heavily from simulation world. Consider a case of wheeled agent touring on an office floor and it crosses over a wet floor that causes its wheel to slip away— resulting in inaccurate odomerty return and that will obviously cause abnormal localization issues. Such events are unpredictable and their definite probability of occurrence is extremely random. Our agent should be proactively capable enough to deal with such unexpected events.

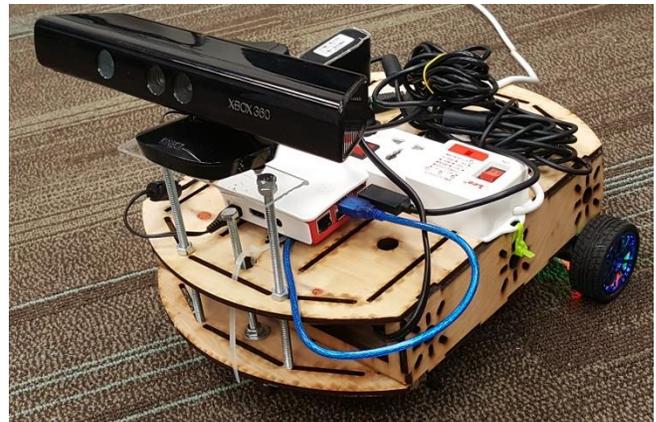

**Fig 1. agent model with (optional) Kinect module**

In this paper, we will analyze a hands-on implementation of A* path finding algorithm on a local hardware model that is entirely hand-crafted (fig.1). The prototyped model is equipped with asynchronous dual DC geared motors with Hall Effect encoders for monitoring odometry. The computational system comprises of Linux (Ubuntu) based Raspberry Pi 3b board and Arduino UNO controller as an auxiliary controller between analog sensors and the on-board-computer (OBC). The agent design also supports Kinect connectivity, if one want to implement simultaneous localization and mapping (SLAM) and online planning & mapping on this design, however, due to time constraints our scope will remain on offline planning (A*) only. We will use Robot Operating System (ROS) environment for creating event nodes for controlling and executing our offline planning system.

Since the agent model is not an off-the-shelf product, one can easily encounter the problems like moving offsets, irregular turns, dumb encoder signals and so forth. We will discuss designing control algorithms to mitigate these physical dynamics problems in the implementation section. Finally, we will test our agent against its simulation results to see how accurately it performs using our feedback and control algorithms.

## 2. BACKGROUND INFORMATION

Wheeled mobile robots mostly operate over a differential drive mechanism. It consists of two motors attached with wheels on a similar axis. Both motor drives are independent of each other's movement. The common access of both motors is known as their center of curvature [1]. The agent cannot move in the direction along axis [2]. Since both motors can move independently, their velocities are also different and we can easily vary the trajectory of movement by varying velocities. Observe that we can maneuver the robot location with angular ($\omega$) as well as linear velocity (V). The velocities of both drives must be synchronized in order to move the agent in the desired path. Also, when the agent is moving in linear (forward/backward) direction, then $\omega$ should be zero. Considering an ideal situation that both drives are synchronized and their $\omega$ and V are approximately same, we can represent angular velocity as:

$$\omega = \frac{V_R - V_L}{length} \quad \therefore length = distance\ between\ wheels$$

Assuming that we don't need diagonal moves, our motion conditions will be: forward, backward, left, right. We can model the locomotion logic using the angular velocity equation above.

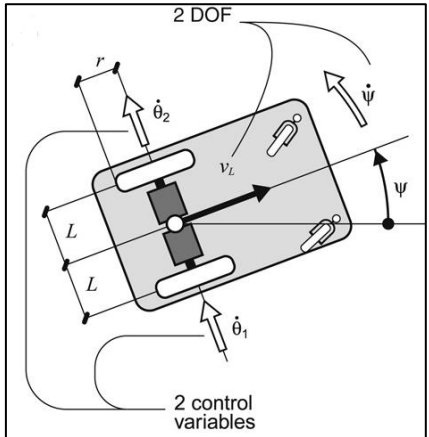

**Fig 2. Differential Kinematic Model** [2]

The cases for our motion will be:

```
          Control case for motor drive directions
1.  if (V_R == V_L)              {FORWARD}
2.  if (-V_R == -V_L)            {BACK}
3.  if (V_L == -V_R)             {RIGHT}
4.  if (V_R == -V_L)             {LEFT}
5.  if (V_R == 0 && V_L == 0)    {STOP}
```

The mathematical model of differential robot tends to be simple, however, the actual differential bot is highly sensitive to minute variations in the velocities of both drives and even a small variation can change the trajectory of a robot. A closed feedback control loop can handle the synchronization by monitoring proportional encoder pulses for each rotation. The control block can be easily implemented within the microcontroller (arduino) environment. The arduino controller can hold multiple interrupt requests at a time and we can hook up interrupt [5]. One benefit of handling control blocks within the microcontroller is that the control signals remain readily available to the agent even if the on-board computer (Raspberry Pi in our case) is power downed. We will describe the interrupt encoder relationship in detail within *System Implementation* topic. Next important parameter for controlling the drive motors is to establish "H-Bridge" path between motor controls and microcontroller. H-bridge is most essential part of the control system since it helps the microcontroller to leash the velocity and rotary directions of the drive motor. Describing the internal architecture of H-Bridge is out of scope for now, only understanding the working principle would suffice for our analysis.

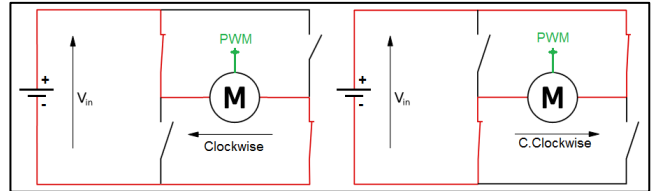

**Fig 3. H-Bridge Working Principle**

The working principle is pretty clear if we observe figure 3. H-Bridge provides switches and a bridge to external power bank for the motors. The switches are handled by our control box. The pulse width modification (PWM) throughputs the velocity/acceleration by an impulse of duty-cycle probed from the controller. Recalling that our control system handles differential drive parameter and H-bridge handlers, now we can step ahead to OBC (Raspberry Pi). The OBC is on the higher level of our system, it runs pathfinding algorithms and returns the required control coefficients to the control unit over USB serial communication node. We use ROS to handle communication node, publishing (TX) and subscribing (RX) nodes between the controller and OBC. Though the literal meaning of publisher and subscriber is deeper than mere TX and RX, but for the sake of simplicity, we will assume publisher and subscriber nodes as TX-RX nodes respectively.

## 3. SYSTEM IMPLEMENTATION

Now as we have a clear idea of system's structure, we can implement control blocks and OBC nodes in order to setup a working pathfinding model. Starting with the control block, we need to create a closed feedback loop for precise calculation.

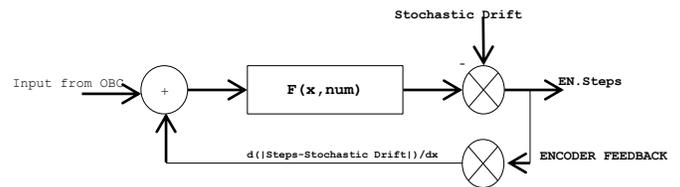

**Fig 4.**

Observing the control loop-consider a single motor drive scenario, the OBC serially sends the number of steps and direction it wants the agent to move. The control block feeds the coefficients to the movement function sitting inside the microcontroller. The function then commands the H-Bridge to make drives moves up to desired steps, however the drives would not exactly follow the number of steps due to stochastic drift. We take the derivative Δsteps and feed it back to the summation node with the input stimulus. This procedure can happen for both motors. Another closed loop will handle the second motor, but the ∆steps will be calculated by adding a comparator at the output of both loops. The comparator then compares `EN.Steps`$_{(R)}$ vs. `EN.Steps`$_{(L)}$, and takes the one with the higher number of steps. Then the comparator output is sent back to either left or right control function (whichever skipped more steps) directly and it sets off the counterbalance. The cascaded feedback function offsets both motor drives equally. And because both outputs are fed to comparator block, therefore positive and negative both offsets are balanced as required. The interesting thing here is to observe that we can easily manage the movements keeping the velocities constant. The trick is that the differential drives are constantly being checked for their feedback steps using hardware interrupt pins of arduino. Also we have simplified each step move by a unit revolution of both wheels. Generally it's not always a good idea to restrict the step size of agent model, but since we are considering only deterministic grids, our system gets more predictable. Moving ahead from low-level controls boxes, we move on to ROS running on our OBC.

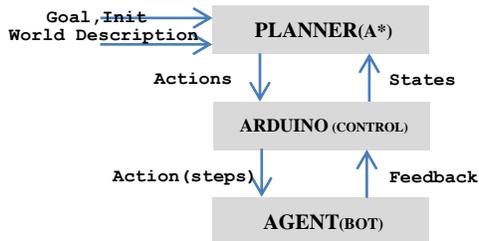

Fig 5

Recalling—refreshing our memories; ROS runs by creating a network of nodes. Each node is connected within a plumbing style structure and has a subscriber and publisher nodes as we discussed earlier. We are using ROS as the master controller over low-level microcontroller based miniature control boxes (fig 5). The ROS environment has a node that publishes the actions our agent needs to take in specific states. Those actions are generated based on the input arguments to the planner. Planner is like a manager who checks up with the occupancy grid and goal location and generates heuristics for finding a traversable path towards goal. We are using A* search algorithm for finding the optimum path towards the goal. The pseudo-code is described below:

1. Place start node in `OpenSet`
2. Calculate the Cost `f(start) = h(start) + g(start)`
3. Remove node `N` from `OpenSet` & put in `ClosedSet`
4. **While `N != GoalNode`**
   - Select a node `N` with `minimum_Cost(OpenSet)`
   - Populate the `NeighborList` for node `N`.
     ➢ If `Neighbor_Node` is in `ClosedSet`
       ▪ Discard that neighbor
     ➢ `Else` put `Neighbor_Node` in `NeighborList`
   - Calculate `f(cost)` for each node in `NeighborList`
   - Add nodes from `NeighborList` to `OpenSet`
     ➢ If `Neighbor_Node` is already in `OpenSet`
       ▪ Update `f(cost)=min[f(new),f(stored)]`
       ▪ Update parent pointer as `N` of that node.
   - Assign `N` as parent to neighbor nodes.
5. Traverse path using parent pointers from `goalNode` to `startNode`

This algorithm is used widespread for pathfinding agents. The reason behind experimenting traditional A* is to test our hardware model against a benchmark algorithm being used for simulation.

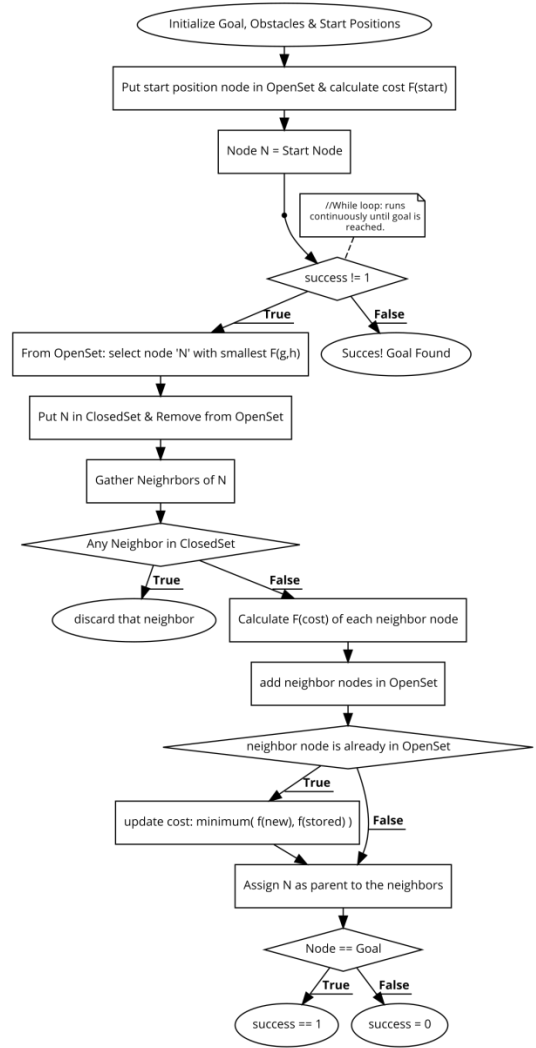

Fig 6

The A* algorithm basically divides the deterministic world grid into nodes and plots the optimal path and then assign heuristics to each node. The heuristics are assigned on basis of the distance from goal to the current node being traversed. There is also a step cost involved in the traversal. The cost can be either uniform for every type of step or it can be more rewarding/penalizing for some steps. The algorithm maintains data structures of the nodes being traversed and also keeps track of the parent node of each successor. The heuristics matrix must be problem specific for map types. Mostly Manhattan distance matrix or Euclidean matrix are seen being implemented. The algorithm is type of informed search problem where you know the goal location, obstacle location and the starting position of the agent. The interesting path of this algorithm is that it always takes you to the shortest or optimal path using cost functions. The cost function comprises of the sum of heuristics value and h(n) and the distance traveled from the starting position. We can represent the cost function as

$$f(n) = h(n) + g(n)$$

During implementation the algorithm checks which of the paths should be expanded based on the estimated of traversal cost + heuristics cost. The two data lists(open & closed) are maintained for node expansion. The un-traversed nodes are populated in open list while the nodes that are either obstacles or are already traversed are put into closed list. After traversing each node, the algorithm again checks for possible expansions and also compare if any of already explored node can be reached with low cost value [6]. If there is any such node in the exploration area, it gets updated with the new low cost value and its new parent node. The data structure keeps on updating until it finally reaches the goal. The path is then back-traced to the starting point by checking the parent nodes of each successor nodes till initial position.

Actually, the planner on a ROS node is exactly same as for a simulation world; the only difference is that it now generates the coordinates of successful simulation episode over its publishing node. Our goal is to get those commands subscribed by the low-level controller. We can easily create a tele-operation environment node to test our hardware model for its response time and behavior upon receiving instruction form tele-operation station. Basically the communication pattern of tele-operation is similar to our planner node, the only difference is that the planner publishes the next state and its action automatically, and the tele-operation requires user input to be sent over publisher. The core part (fig 5) of the whole experiment is making the agent follow the actions without any glitches on the real-world terrain. The system architecture of agent model can be sub-divided into three stations (fig 7)—host, Raspberry Pi or OBC, and Arduino or Control Box. Host station can be a computer or SoC based station, capable of running regression models and maintaining a communication protocol within active ROS environment.

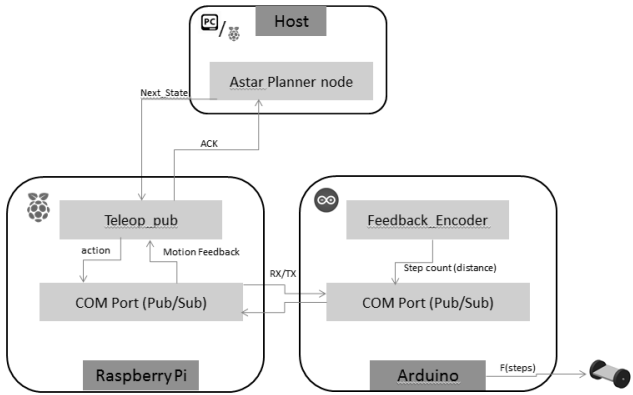

**Fig 7**

We can also create a wireless Ad hoc mesh between each stations but each of the node must be connected on the same network, sharing a common port and access point. The host station broadcasts the next state, our agent need to take. The state argument is generated after receiving acknowledgement form the middle node (Pi node). The Raspberry Pi node is crucial for other node ends because it works as host as well as source simultaneously for other nodes. It also handles the publisher that generates main action (forward, back, left, right) and sends to the microcontroller. And finally the arduino station work as low level ROS node, talking with the actual real world, and acknowledging back the correct or incorrect action taken by our real-world agent.

## 4. RESULTS

The overall study emphasize on practical implementation of the path finding algorithm and observing its behavior in real world grid. Following significant parameters were used for modeling the physical system.

| Parameters | Value |
| --- | --- |
| Baud Rate | 115200 |
| Encoders | Hall Effect Dual Encoders / wheel |
| Encoder Pulses | 440 pulses per revolution |
| Distance / Step | 8 inches per revolution (1 step = 1 rev) |
| Distance Error / Step | +0.36 inches |
| Encoder Offset / Step | ~ +66 to +190 |
| Video Demonstration | https://www.youtube.com/watch?v=bz_q9yjiNwg |

We have evaluated the system against ideal simulation world and have found that the ground moves of our empirical hardware model are eventually satisfying in terms of per-step calculations, moving from one node to another and performing exact actions as required. However the non-synchronization of motors are prone to error with simple $1^{st}$ order feedback loop which we used in our model.

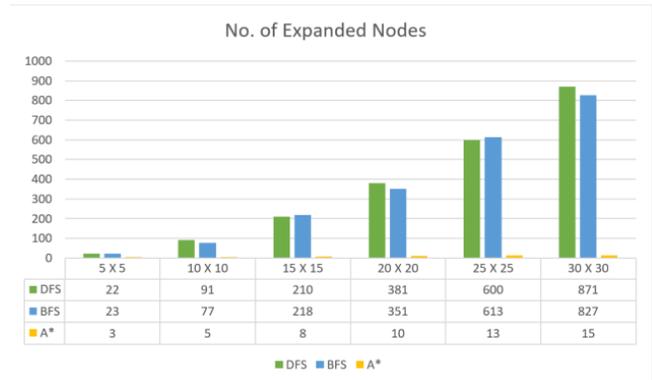

**Fig 8**

We start by an overall comparison of the different variants of search algorithms for finding the path to help the agent reach the goal. Fig 8 Shows the comparison of Depth First Search and Breadth First Search vs A * on a grid world which ranges from dimensions of 5 X 5 to 30 X 30. It was observed that DFS and BFS expanded almost every cell in the grid world while looking for the goal state. The time the agent took to explore the grid world using DFS and BFS search algorithm differed by a relatively small amount of time than A * search algorithm, but as the size of the grid world increased, A * was found to be the most efficient. In some cases, BFS performed well than DFS but A * was always better, both in terms of space complexity and time complexity. In A * algorithm, which is a family of algorithm, where the choice of a particular function h (heuristic) selects a particular algorithm from the family. The function h can be used to tailor A* for particular applications. In our demo of pathfinding robot with obstacles in between, the robot looks for the goal state and directly approaches towards it. It goes for the greedy approach and if it finds an obstacle in between, it still expands the node and then looks for the shortest path by trying to explore the minimum nodes thereafter. We use different heuristics (Euclidean and Manhattan) to find the distance from start state to goal state

When using the Euclidean space, we increase h(n) from 0 to $\sqrt{x^2 + y^2}$ (x and y are the magnitudes of the differences in the x and y co - ordinates). The algorithm will still find the shortest path, but would do so by expanding less number of nodes. In this method, the computational effort is more since the Euclidean distance formula is complex. Now suppose we want to reduce our computational effort we choose the heuristic h(n) to be (x + y)/2 where it does not involve complex computations and is still regarded as admissible heuristic. If we want to reduce our computational effort still further, we use the Manhattan distance, to calculate the heuristic, which is calculated by simply adding x and y where x and y are the magnitudes of the differences in the x and y co – ordinates. This reduces the computational effort and is still an admissible heuristic. In some maps where a robot finds an obstacle while approaching the goal, it tries to expand on all the sides and reach the goal. Here, the computational effort increases if the nodes it is expanding have the same f values. So to avoid this situation and save time, we use tie breaking where, when the robot encounters such a situation, it will take any one of the path and continue to expand on that path instead increasing the efficiency

## 5. CONCLUSION

We have presented a detailed walkthrough for designing hardware based real-world robot agent. The implemented model is designed to execute popular A* algorithm in the real-world grid system. Factors like robot kinematics, drive coordination, offset handling and communication were under evaluation. The final outcome of the procedure followed in this document leads to a near accurate wheeled robot model. While development we also faced failures and extreme challenges due to exogenous events during runtime. Such events include wheel slipping, taking actions in wrong directions, highly asynchronous motor behavior, and fluctuating power supplies that impact the sensor data drastically.

These problems were addressed using the closed loop feedback control techniques and by using pre-defined environments like ROS that compensate errors significantly.

For future research ideas, we would recommend implementing SLAM models and would recommend using advanced localization techniques for online planning and mapping. Also localization using non-parametric filters would highly aid towards robustness of the proposed hardware model.